\documentclass[10pt, conference, compsocconf]{IEEEtran}

\IEEEoverridecommandlockouts                     

\usepackage{cite}
\usepackage{amsmath,amssymb,amsfonts}
\usepackage{graphicx}
\usepackage{textcomp}
\usepackage{xcolor}
\usepackage{caption}
\usepackage{algorithm}
\usepackage{algpseudocode}
\usepackage{subcaption}
\usepackage{lipsum}
\usepackage{float}
\graphicspath{{images/}}

\def\BibTeX{{\rm B\kern-.05em{\sc i\kern-.025em b}\kern-.08em
    T\kern-.1667em\lower.7ex\hbox{E}\kern-.125emX}}
\begin{document}

\title{A Deep Learning-Based Autonomous Robot Manipulator for Sorting Application}


\author{Hoang-Dung Bui, Hai Nguyen, Hung Manh La, Shuai Li
\thanks{This material is based upon work supported by the National Aeronautics and Space Administration (NASA) Grant No. NNX15AI02H issued through the NVSGC-RI program under sub-awards No. 18-94  and 20-16. This work is also partially supported   by the U.S. National Science Foundation (NSF) under grants NSF-CAREER: 1846513 and NSF-PFI-TT: 1919127.  The views, opinions, findings and conclusions reflected in this publication are solely those of the authors and do not represent the official policy or position of the NASA and NSF.}
\thanks{The first two authors have made equal contributions to this work.}
\thanks{H.D. Bui, H. Nguyen and H. La are with the Advanced Robotics and Automation (ARA) Laboratory, Department of Computer Science and Engineering, University of Nevada, Reno, NV 89557, USA.}%
\thanks{S. Li is with the College of Engineering,
Swansea University, Fabian Way, Swansea, SA1 8EN, Wales, UK.}
\thanks{ Corresponding author: Hung La ({\tt\small e-mail: hla@unr.edu})}
\thanks{Video demonstration of this implementation is available at https://www.youtube.com/watch?v=ujorCGl5Ieo.}
\thanks{Source code is available at the ARA lab's github: https://github.com/aralab-unr/GraspInPointCloud.}
}
\maketitle
\begin{abstract}
Robot manipulation and grasping mechanisms have received considerable attention in the recent past, leading to development of wide-range of industrial applications. This paper proposes the development of an autonomous robotic grasping system for object sorting application. RGB-D data is used by the robot for performing object detection, pose estimation, trajectory generation and object sorting tasks.
The proposed approach can also handle grasping on certain objects chosen by users. Trained convolutional neural networks are used to perform object detection and determine the corresponding point cloud cluster of the object to be grasped. From the selected point cloud data, a grasp generator algorithm outputs potential grasps. A grasp filter then scores these potential grasps, and the highest-scored grasp will be chosen to execute on a real robot. A motion planner will generate collision-free trajectories to execute the chosen grasp. The experiments on AUBO robotic manipulator show the potentials of the proposed approach in the context of autonomous object sorting with robust and fast sorting performance. 
\end{abstract}


\section{Introduction}

Object sorting has numerous applications in a diverse range of environments and contexts, ranging from household and industrial settings to agriculture and pharmaceutical industries. However, objects sorting tasks performed by human beings are tedious and error-prone in nature, especially over extended periods of time. In order to improve on the deficiencies of human-based sorting applications, the use of autonomous or semi-autonomous robots has been proposed in recent studies \cite{gupta2012using, zeng2018robotic, guerin2018unsupervised, mahler2019learning}. In \cite{gupta2012using}, Gupta at al. proposed a framework, which sorted the simple Duplo bricks by size and color by using depth data to determine the grasp pose for the bricks. However, its application was limited to very light objects with simple geometries. In \cite{zeng2018robotic}, Zeng at al. proposed an robotic pick-and-place system, which was able to grasp and recognize objects in cluttered environment. In this algorithm, the object recognizing happened after grasping, and it seemed suitable for the cleaning task than object's sorting. In \cite{guerin2018unsupervised}, Guerin et al. applied an unsupervised deep neural network for a robotic manipulator to classify the object based on feature extraction and standard clustering algorithm. It sorted well the objects with similar geometries, however, it failed to classify the same type objects with different dimensions. 

To work in complex environment \cite{JinTIE2017,NguyenIRC2019}, an autonomous sorting machine should have the following capabilities: (i) detection and classification of objects with different shapes, sizes and physical properties, (ii) optimal object grasping, and (iii) trajectory generation and motion planning within the 3D environment. In this paper, we present our development of an autonomous object sorting system using robotic grasping mechanism. By using deep neural networks, the system is able to detect multiple types of objects, select an optimal grasping object, and its optimal grasp pose, perform the grasp action on a real robotic manipulator. Particularly, the contribution of this paper are:
\begin{itemize}
    \item development of  a complete integration system of robotic sorting manipulator,
    \item combination of two convolutional neural networks (CNNs) to be able to process RGB-D data to do both duties of object detection and object grasping.
\end{itemize}
Combining with \textit{Trajopt} \cite{schulman2013finding} motion planner, the experiments on AUBO robotic manipulator show the potentials of the proposed approach in the context of autonomous object sorting with robust and fast sorting performance.

This paper has been divided into five sections. In section II, state-of-the-art, related to object detection, grasping mechanisms and robotic motion planning for pick-and-place operations, will be presented. Section III will discuss the salient features of the proposed method for development of object sorting system using autonomous robotic manipulator and Deep CNNs. Section IV will discuss the different aspects of the experimentation and associated results. Section V will conclude the research findings and provide recommendations for future research in relevant research area.

\begin{figure*}[ht]
\centering
\includegraphics[width=0.8\textwidth]{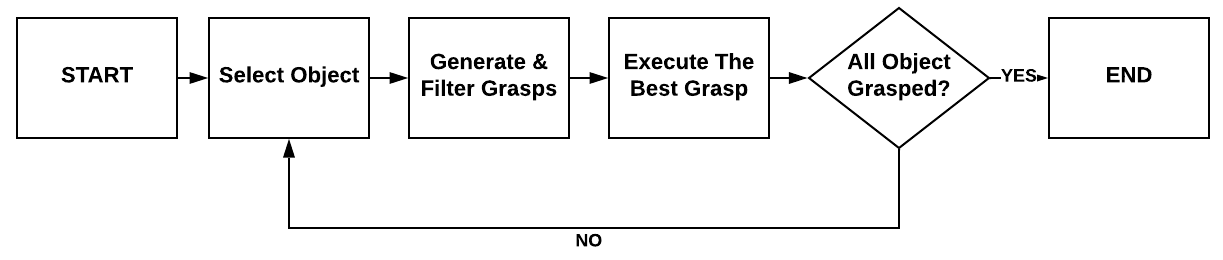}
\caption{Proposed approach.}
\label{fig:proposed_approach}
\end{figure*}

\section{Related Work}
This section discusses the state-of-the-art technologies in Object Detection and Object Grasping. 
\newline
\newline
\textbf{Object Detection}. 
The sorting system runs in real-time manners and performs both detection and grasping work. Thus, it requires that the used algorithms should meet the real-time running, and provide the coordinates of detected objects. Applying deep neural network in object detection has improved in term of accuracy and real-time processing even with limited computational resources \cite{SehgalIRC2019}. 
The CNNs in \cite{redmon2018yolov3, liu2016ssd}  are region proposal based framework, which mapped straightly from image pixels to bounding box coordinates and class probabilities, thus reduce time expense for shared convolution parameters.
Liu et al. \cite{liu2016ssd} proposed a Single Shot MultiBox Detector (SSD), which takes advantage of a set of default anchor boxes with different aspect ratios and scales to  discretize the output space of bounding  boxes. To handle objects with various sizes, the CNN fuses predictions from multiple feature maps with different resolutions. 
Given the VGG16 \cite{simonyan2014very} backbone architecture, SSD adds several feature layers to the end of the network to predict the offsets to default boxes with different scales and aspect ratios and their associated confidences. The network is trained with a weighted sum of localization loss and confidence loss. SSD runs at 59 frame per second (FPS) with 28.1 mean Average Precision (mAP), however, it does not handle well with small objects.

Redmon et al. \cite{redmon2018yolov3} proposed a novel framework called YOLO to predict both confidences for multiple categories and bounding boxes. The YOLOv3 consists of 53 conv layers of which some conv layers construct ensembles of inception modules with 1x1 reduction layers followed by 3 x 3 conv layers. It is able to process an image in 22 \textit{ms} at 28.2 mAP and classify more than 80 object classes. With real-time operation capabilities, efficient performance and versatility for object detection, YOLOv3 was a good option  for our proposed system. Even though, we need to modify the output of YOLO to get the centroid of each object selection.
\newline
\newline
\textbf{Grasp Detection}. 
The use of point cloud data with CNN has been used to provide a reliable object grasp pose with varying pose and finger gripper configurations \cite{NguyenICDL-EpiRob19, NguyenIRC2019}.

In \cite{redmon2015real, lenz2015deep}, the authors tried to find good grasp poses using RGB-D image frames with single-state regression to obtain bounding boxes containing target objects. The algorithm in \cite{redmon2015real} reached great speed up to 3 FPS on a GPU and high accuracy of 90.0$\%$ on image-wise split. However, it only provided the pose in 2D, and lacked of pose orientation in depth direction. The algorithms in \cite{lenz2015deep} considered the depth data and outputted reliable grasp poses, which can be graspable for finger gripper configuration. The drawback is that it is is not a real-time application due to the processing time.

In \cite{ten2018using, gualtieri2016high}, the authors used another neural network called Grasp Pose Detection (GPD) to improve the quality of detected grasp poses. The input of the CNN was the object point cloud data, which processes the local geometry and graspable surfaces of the objects. To speed up the processing time, the authors proposed two new representation of grasp candidates and trained the CNN with large online depth datasets obtained from idealized CAD model. Their approach, however, failed to address the difficulty of distinguishing between two adjacent objects as the algorithm avoids point cloud segmentation. As a result, there is also no direct way to grasp a specific object of interest. If the problem of point cloud segmentation is solved, it means that a grasp pose is generated for a single standing object, this approach is suitable with our work and provide a reliable grasp for specific type of object.

\begin{figure*}[ht]
\centering
\includegraphics[width=0.7\textwidth]{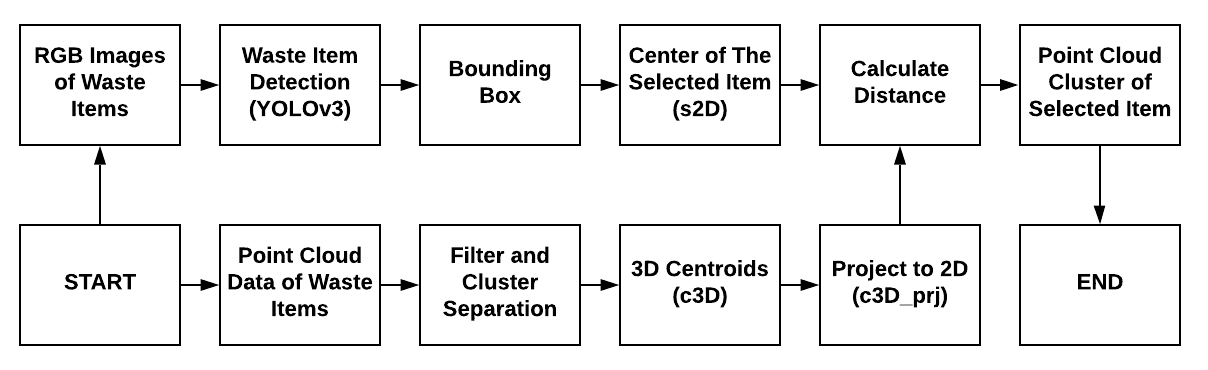}
\caption{Flowchart of selecting an object to grasp.}
\label{fig:flow1}
\end{figure*}

\section{The Proposed Method}
The proposed approach included three stages as shown in Figure \ref{fig:proposed_approach}:

\begin{itemize}
    \item \textit{Object selection}: In this stage, the robot system has to select one object out of a multitude of different objects present within a  given environment. The priority for object selection can be specified by the users, in terms of the following criteria: proximity, object class, physical properties. If there are multiple objects in the same type, the system needs criteria to score them and grasp each item in a predefined order. This step uses the first CNN to detect, select then output the point cloud cluster of the selected object, which is sent to the second deep network to generate the grasps.
    \item \textit{Grasp Pose Selection}: In this step, the input is the selected object's point cloud data, which is used to estimate and generate grasp poses on the object. The second CNN generates a number of different poses, then the most suitable grasping pose is selected. This requires a filter applied on candidate grasp poses, which is scored and the highest scored pose is selected and given as output to the next step.
    \item \textit{Object Grasping and Sorting}: A motion planner is used to generate a trajectory that helps the robot's gripper to reach the desired grasp pose for grasping target object. It is important for the generated robot trajectory to be collision-free. The motion planner, therefore, needs a model of the environment so that it can check for collisions when generating trajectories.
\end{itemize}

\subsection{Object Selection}
It is challenging to perform object detection directly using point cloud data as input. Several visual cues such as colors or shapes, that are normally used to recognize an object can be affected. To overcome the challenge, color images and point cloud data are combined to make the decision. The steps for object selection are illustrated in Figure \ref{fig:flow1}. RGB images of items are given as input to an image detector to put a bounding box around the selected object.

\begin{figure}[ht]
\centering
\includegraphics[scale=0.14]{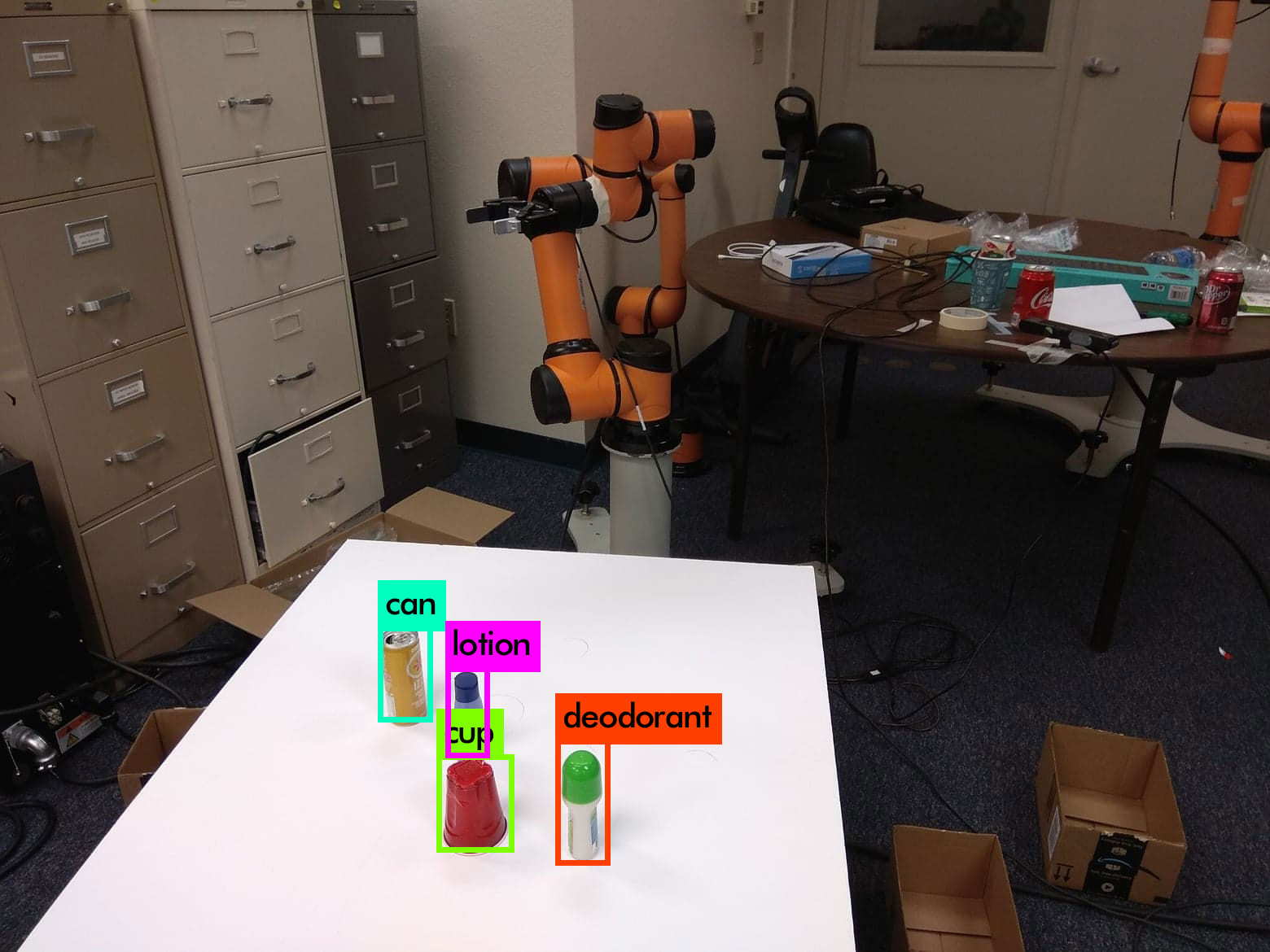}
\caption{Objects detection by using YOLOv3 network.}
\label{fig:YOLO}
\end{figure}

YOLOv3 provides the performance metrics for object detection and selection, which is used to score each item in the same class in case multiple items are available. The object with the highest score is given as output to the next process. The center of the bounding box $s_{2D}$ (as shown in Figure \ref{fig:YOLO}) is calculated.

\begin{figure}[ht]
\centering
\includegraphics[scale=0.45]{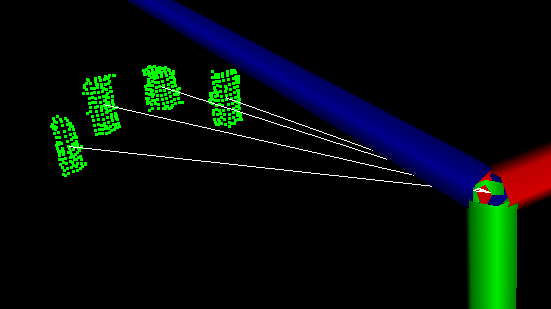}
\caption{Corresponding clusters of objects in Figure \ref{fig:select_object} are separated. The red, green, and blue bars are $xyz$ axes of the camera's depth coordinate frame.}
\label{fig:cluster}
\end{figure}

In another process, point cloud data of different objects is filtered to remove noise, outliers, and reduce the amount of data by using the following: statistical filter, a voxel filter, and a working space filter. A workspace filter is used to remove the data points that do not belong to a predefined workspace. At this step, plane segmentation and target extraction is used to separate the items as shown in Figure \ref{fig:cluster}.

\begin{figure}[ht]
\centering
\includegraphics[scale=0.35]{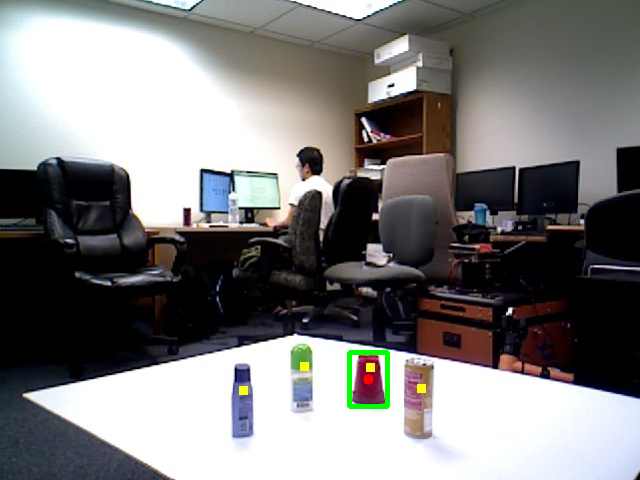}
\caption{\label{fig:select_object}Four objects to be selected. The cup is chosen to be grasped.}
\end{figure}

\begin{figure}[ht]
\centering
\includegraphics[width=1.0\linewidth]{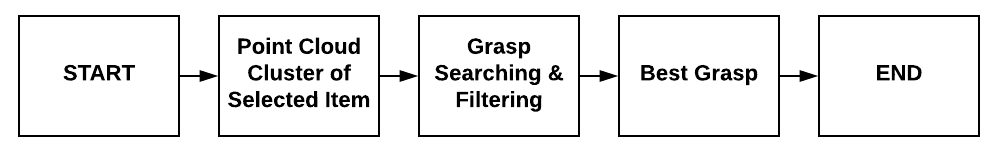}
\caption{Grasp generation \& filtering on the selected item.}
\label{fig:grasp generating}
\end{figure}

Also shown in Figure \ref{fig:cluster}, the 3D centroid coordinates of each object $c_{3D}^i$ are calculated, where $i$ is index for the object cluster $i$. These centroids are then projected to get 2D points $c_{3Dp}^i$ in the image coordinate (using the camera calibration parameters). At the last step, the Euclidean distances between $s_{2D}$ and $c_{3Dp}^i$ are calculated, and the smallest distance is used to determine the corresponding cluster belonging to the selected object:

\begin{itemize}
    \item Calculate the distance $d_i$ for cluster $i$: 
    \begin{equation*}
        d_i^2 = (s_{2D_x} -c_{3Dp_x}^i)^2 + (s_{2D_y} -c_{3Dp_y}^i)^2.
    \end{equation*}
    \item Compare and pick the cluster with the smallest distance.
\end{itemize}
All the coordinates in the formula are calculated from the camera base frame.

To illustrate the process, for instance, a cup is the target object to be grasped by the manipulator. The process to determine the corresponding cluster is shown in Figure \ref{fig:select_object}. The yellow squares are $c_{3Dp}$ points, and the red circle is $s_{2D}$, which is calculated from the coordinates of the bounding box of the cup returned by the trained YOLOv3 network. The nearest yellow square to the red circle determines the third cluster from the left (as shown in Figure \ref{fig:cluster}) belongs to the object to be grasped (the cup). After that, the grasp detection algorithm can be performed on this cluster.

In Figure \ref{fig:cluster}, four set of green points corresponding to the point cloud data of four objects shown in Figure \ref{fig:select_object}. The red, green, and blue bars denote $x$, $y$, and $z$ axes of camera's depth coordinate frame.

\subsection{Grasp Generation and Filtering}
The flowchart for grasp posture estimation and generation is illustrated in Figure \ref{fig:grasp generating}. After having the cluster of the grasped object, the algorithm for Grasp Pose Detection \cite{gualtieri2016high} is performed. Each point in the cloud is associated with a single viewpoint, from which the point is captured. The algorithm also considers the geometric parameters of the robot gripper and a subset of points $C_G$ belonging to target objects. 

\begin{figure}[ht]
\centering
\includegraphics[scale=0.15]{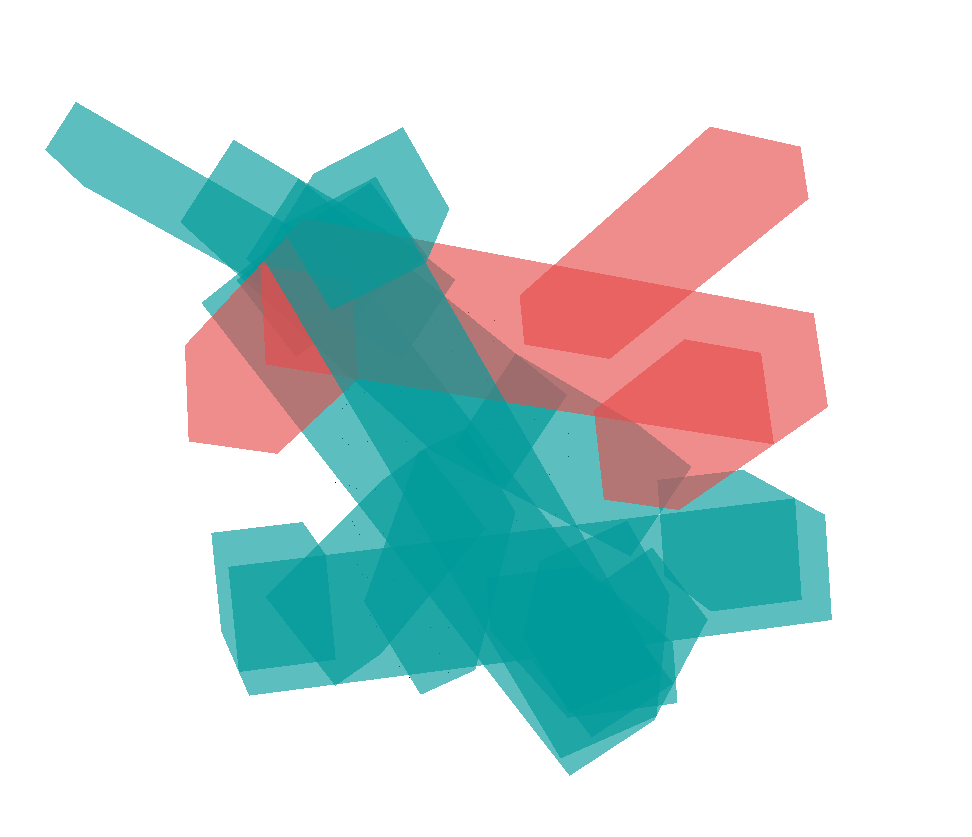}
\caption{Eight highest scored poses to grasp an object and the best pose is in red. This posture points toward the object and is similar to the current gripper's pose. Moreover, the contacting point of grasp pose in on the upper part of the target object.}
\label{fig:filtered_grasp_pose}
\end{figure}

To have the subset of points that belong to objects of interest $C_G$, those central points calculated earlier from the labeled bounding boxes and spheres of points around them are used. With point cloud data and centroid points as inputs, the algorithm samples points uniformly around $C_G$. After that, the algorithm calculates the surface normal and an axis of major principal curvature of the object surface. Potential grasp pose candidates are generated at regular orientations orthogonal to the curvature axis. These grasp poses are then pushed forward until the fingers make contact with the point cloud. The grasps poses without any data points between the fingers are discarded. The remaining grasp poses are given as input to a four-layered CNN for pose classification between viable and non-viable grasping poses. At this step, the algorithm is used for scoring grasp poses to pick the grasp pose with highest score. The algorithm gives a high score for grasps that are at the upper part of the object (higher chance of successful grasps). It also considers the gripper orientations and poses if they are similar to the current pose of the gripper or points towards the robot position. This helps to minimize the robot's movements. In Figure \ref{fig:filtered_grasp_pose}, the best pose to grasp is the red one.

\subsection{Grasp Execution}
\begin{figure}[ht]
\centerline{\includegraphics[width=1.0\linewidth]{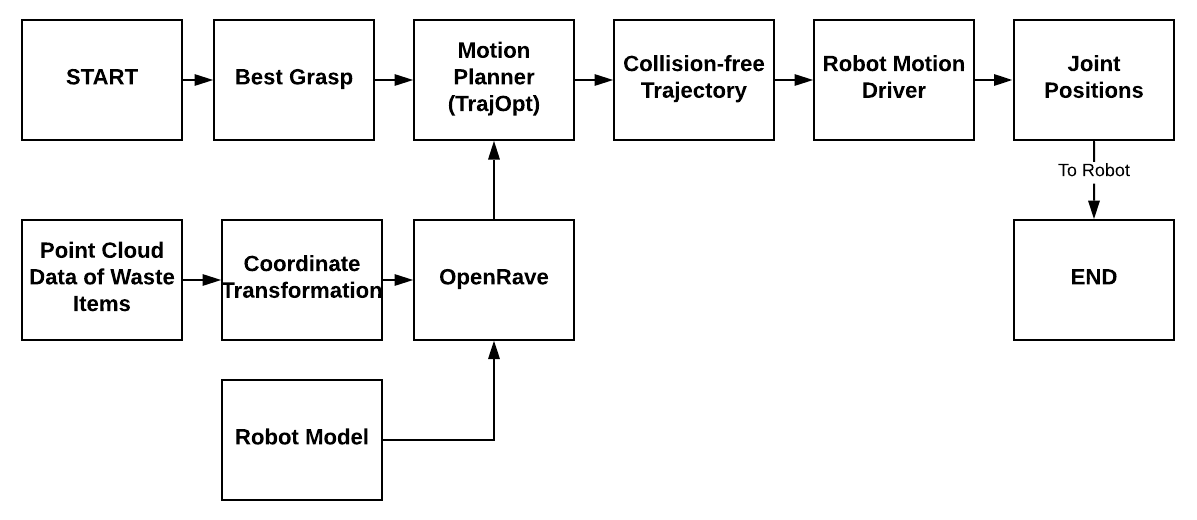}}
\caption{Flowchart of executing the chosen grasp.}
\label{fig:grasp_executing}
\end{figure}

To execute the chosen grasp on a real robot, there is a need to generate a collision-free trajectory, which transforms the current gripper's pose to the desired grasp pose. To check for collisions, \textit{TrajOpt} \cite{schulman2013finding} needs an accurate simulated model of the environment, which includes the robot model and the point cloud data of objects in the robot base coordinate. The execution of the chosen grasp was illustrated in Figure \ref{fig:grasp_executing}.

\begin{figure}[ht]
\centering
\includegraphics[width=0.3\textwidth]{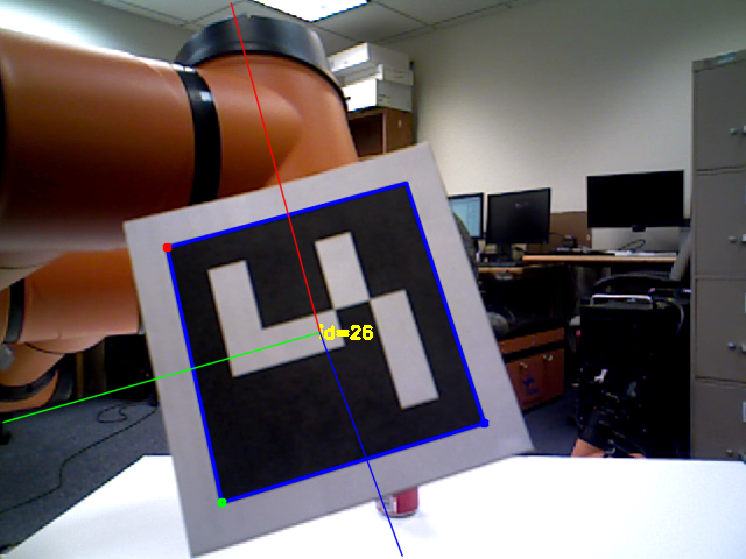}
\includegraphics[width=0.3\textwidth]{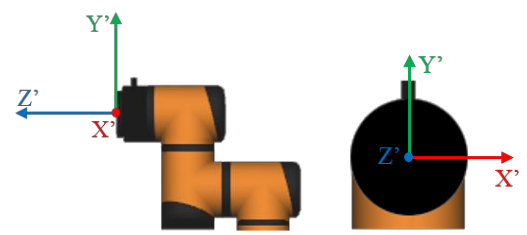}
\includegraphics[width=0.3\textwidth]{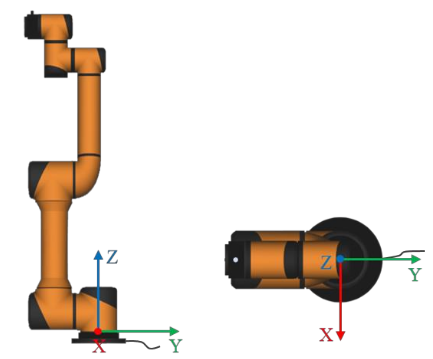}
\caption{Coordinates: top - Aruco tag's coordinate $C_{tag}$ in the camera coordinate $C_{cam}$; middle - end-effector coordinate $C_e$, bottom - robot base coordinate $C_{base}$. Color: $X,Y,Z$ = red, green, blue.}
\label{fig:coordinates}
\end{figure}

\begin{figure}[ht]
\centering
\includegraphics[width=0.5\textwidth]{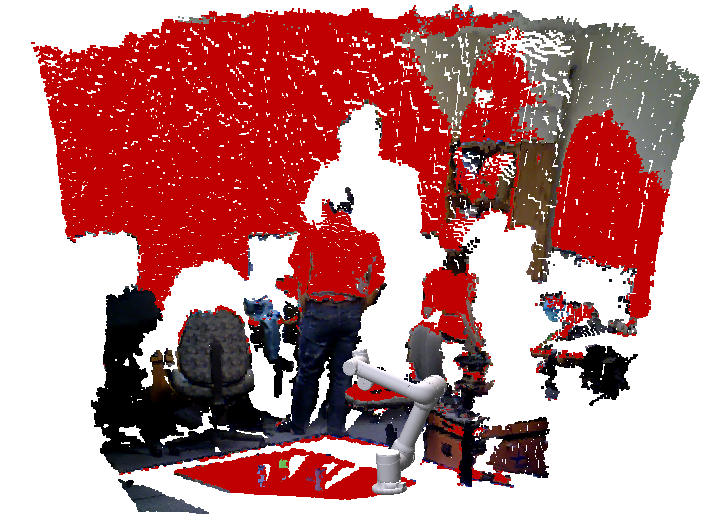}
\caption{Simulated environment in OpenRave.}
\label{fig:All}
\end{figure}

As all the objects that the camera sees are in the camera coordinate, there is a need to calculate a transformation matrix from this coordinate to the robot base coordinate. The Aruco tags \cite{garrido2014automatic} were used to calculate the transformation matrix through the end-effector coordinate of the robot. These coordinates are defined in Figure \ref{fig:coordinates}.
\begin{itemize}
    \item An Aruco tag is attached on the end-effector of the robot (the gripper is temporarily removed) so that the transformation matrix between the tag's coordinate $C_{tag}$ and the end-effector coordinate $C_e$ is known.
    \item The transformation matrix between $C_{tag}$ and the camera coordinate $C_{cam}$ can be computed using any Aruco tag software package. We use aruco\_ros, and the detection result is shown in Figure \ref{fig:coordinates}.
    \item The transformation matrix between the $C_e$ and $C_{base}$ is provided by the forward kinematics of the robot.
\end{itemize}
The homogeneous transformation matrix between $C_{cam}$ and $C_{base}$ is the multiplication of the above matrices:
\begin{equation*}
    M^{base}_{cam} = M^{base}_{e} M^{e}_{tag} M^{tag}_{cam}.
\end{equation*}
Using $M^{base}_{cam}$, the point cloud data in $C_{cam}$ can be easily converted to $C_{base}$. Using this matrix, the point cloud data from the camera's coordinate can be transformed and displayed them in the robot base coordinate using OpenRave \cite{diankov2008openrave} as shown in Figure \ref{fig:All}.

After transformation has been performed to $C_{base}$ that includes the robot, the objects, and the desired grasp pose; TrajOpt will generate a collision-free trajectory. The trajectory is a series of 6-joint position tuples, which can then be successfully performed on the actual robot.

\begin{figure}[ht]
\centering
\includegraphics[width=0.47\textwidth]{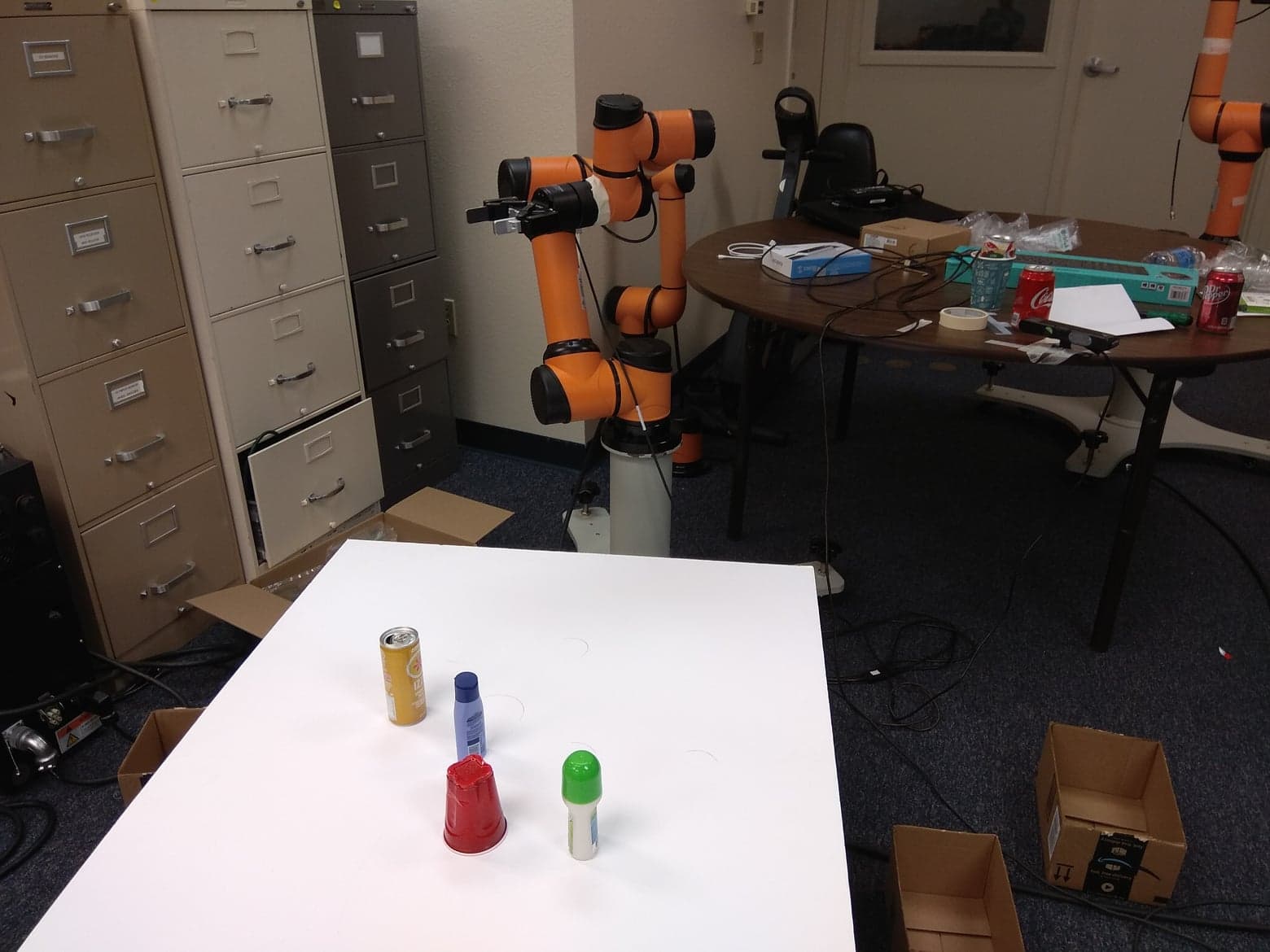}
\caption{\label{fig:Setup}Hardware setup.}
\end{figure}

\begin{figure}[ht]
\centering
\includegraphics[scale=0.16]{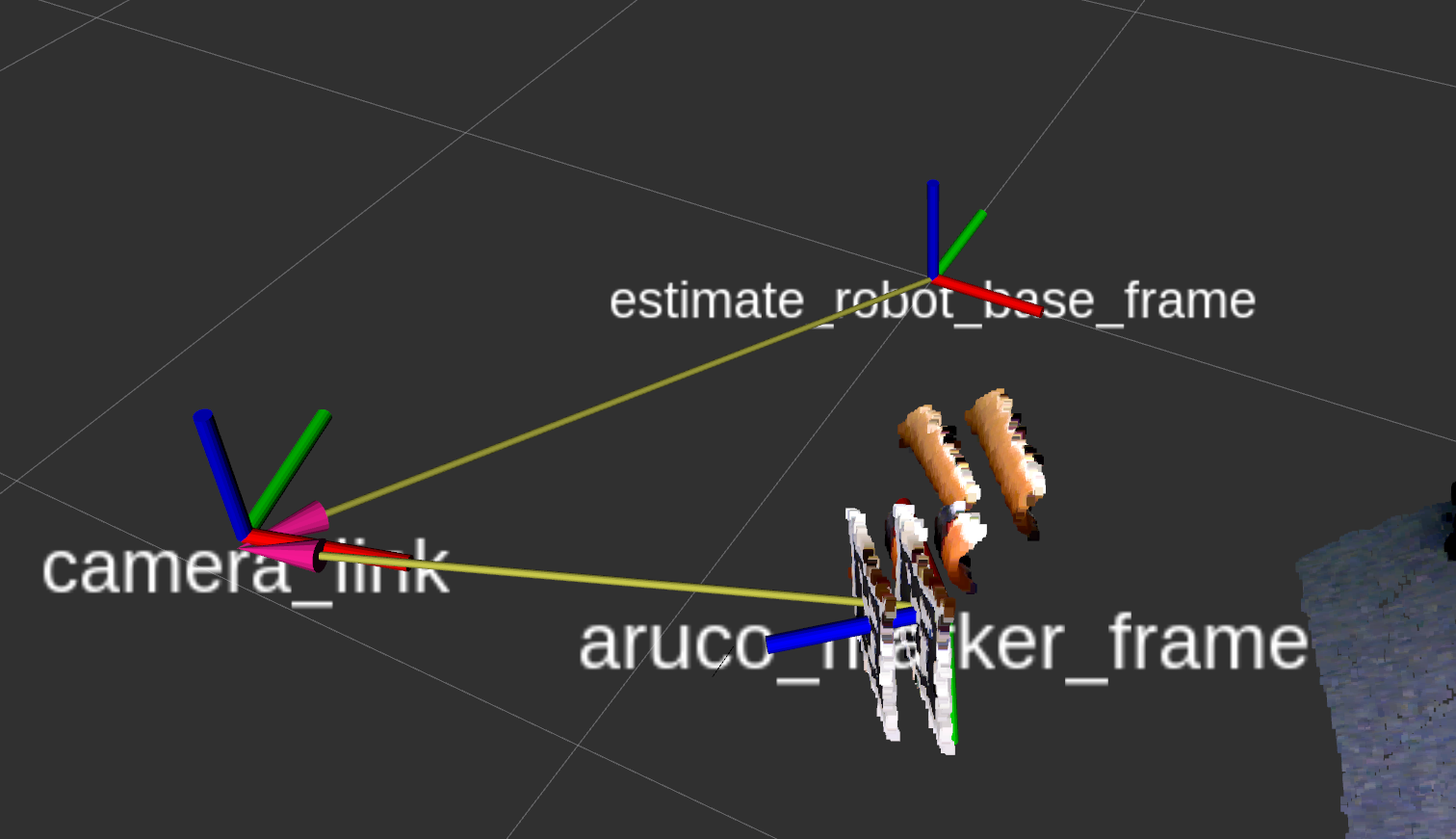}
\caption{\label{fig:depth-error}Mitigated depth error after a calibration.}
\end{figure}

\section{Experiment}
\subsection{Hardware}
An AUBO-i5 robot from AuboRobotics 6-DOF with payload of 5 kg is used in the practical experimentation. The robot is controlled using a Python driver provided by the manufacturer. A Robotiq 2-finger 85 adaptive gripper is mounted on the arm. The gripper is position-controlled with positional and force feedback and adjustable gripping force between 20N and 235N. For the vision system, Asus Xtion Pro RGB-D camera has been used. The camera outputs 640x480 RGB images along with point cloud data. The camera is fixed at a position on a table during the experiment. The whole setup is illustrated in Figure \ref{fig:Setup}.

\begin{figure*}[ht]
    \centering
    \begin{subfigure}[b]{0.2\textwidth}
        \centering
        \includegraphics[height=3cm]{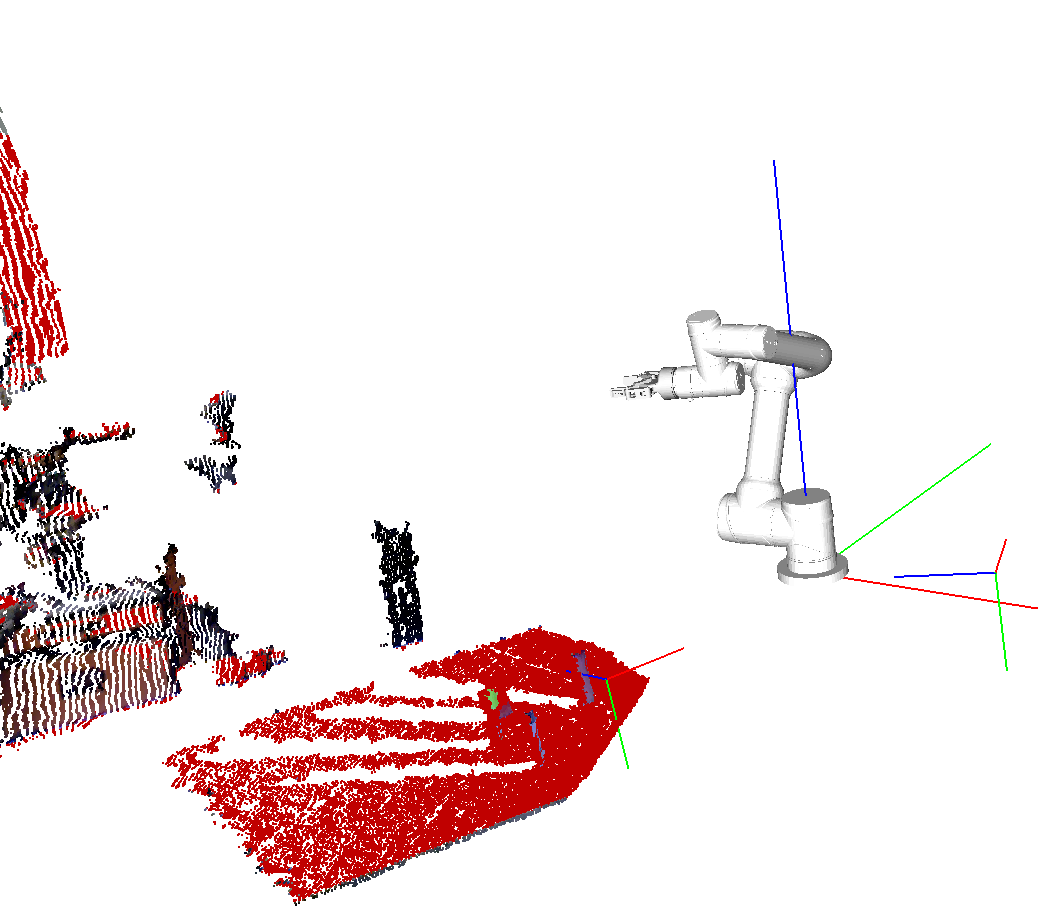}
        \caption{Robot \& scene loaded.}
    \end{subfigure}
    \hspace{\fill}
    \begin{subfigure}[b]{0.2\textwidth}  
        \centering 
        \includegraphics[height=3cm]{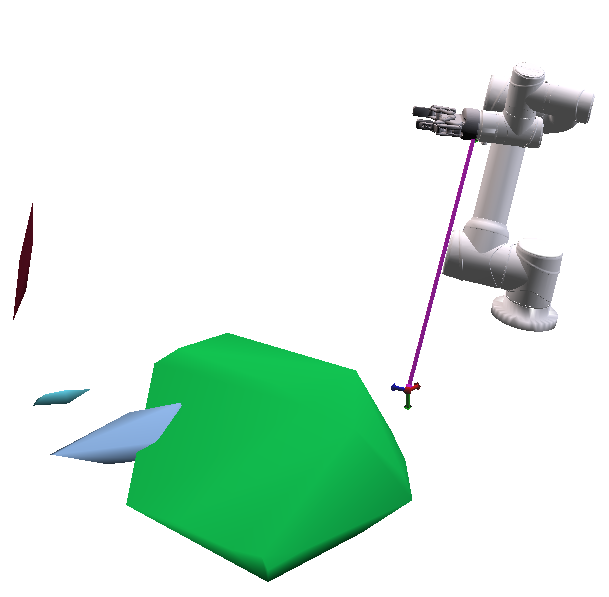}
        \caption{Planning is started.}
    \end{subfigure}
    \hspace{\fill}
    \begin{subfigure}[b]{0.2\textwidth}  
        \centering 
        \includegraphics[height=3cm]{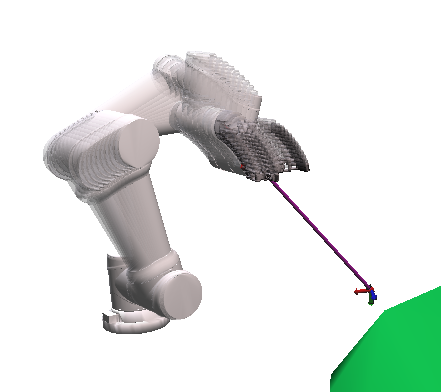}
        \caption{Building trajectories.}
    \end{subfigure}
    \hspace{\fill}
    \begin{subfigure}[b]{0.2\textwidth}   
        \centering 
        \includegraphics[height=3cm]{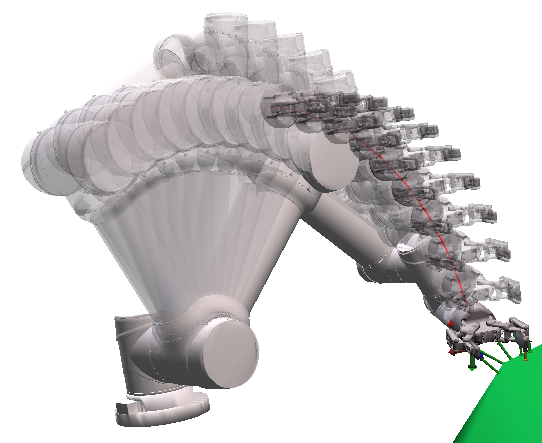}
        \caption{Path planning finished.}
    \end{subfigure}
	\caption{Planning for one grasp pose by OpenRave. First (a), the robot model, the point cloud data, and the desired grasp pose of the scene are loaded. The robot coordinate and the desired grasp pose are plotted with $X$ axis (red), $Y$ axis (green), $Z$ axis (blue). Next (b), the point cloud data is simplified by approximated meshes for collision checking, starting the path planning. (c) Trajectories is then built. (d) The trajectory (red) is generated successfully without any collision.}
    \label{fig:planning}
\end{figure*}

\begin{figure}[ht]
    \centering
    \begin{subfigure}[b]{0.2\textwidth}
        \centering
        \includegraphics[height=2.5cm]{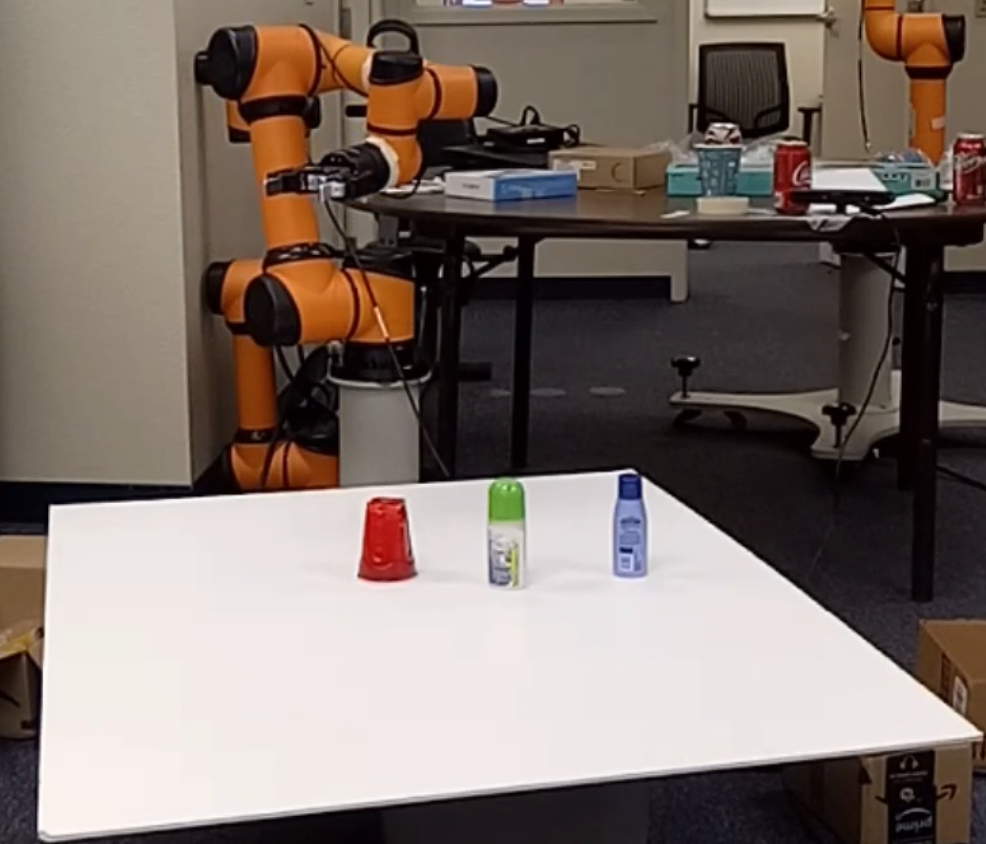}
    \end{subfigure}
    \hspace{\fill}
    \begin{subfigure}[b]{0.2\textwidth}  
        \centering 
        \includegraphics[height=2.5cm]{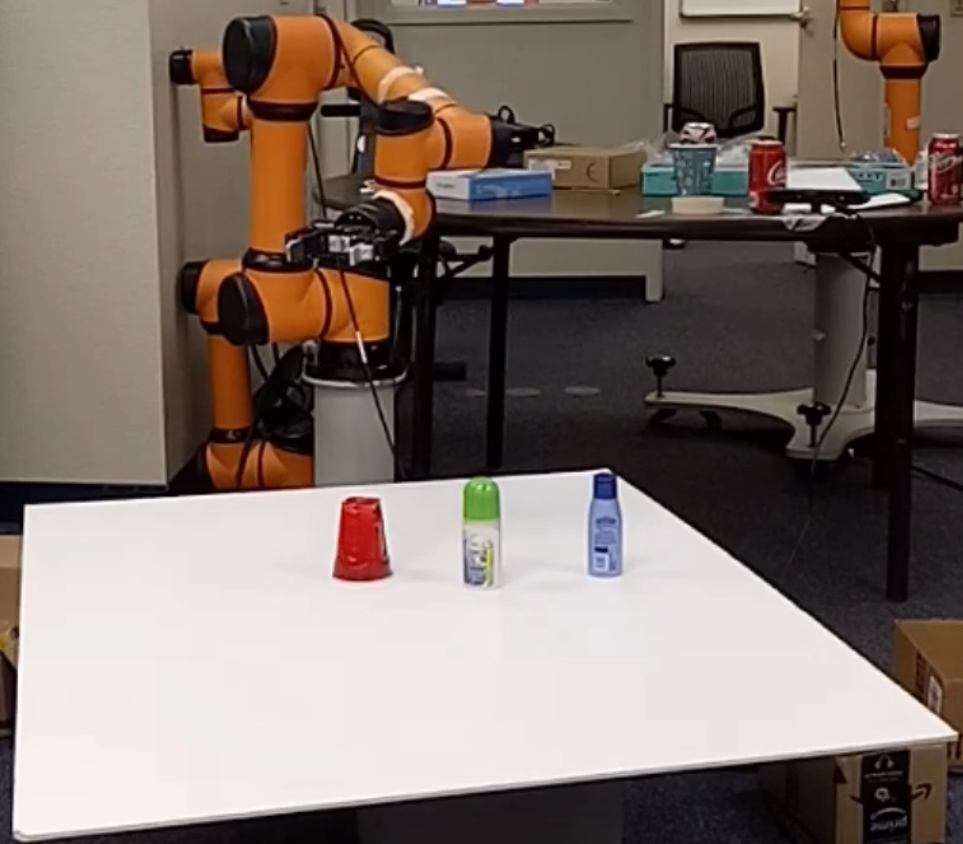}
    \end{subfigure}
    \hspace{\fill}
    \begin{subfigure}[b]{0.2\textwidth}   
        \centering 
        \includegraphics[height=2.5cm]{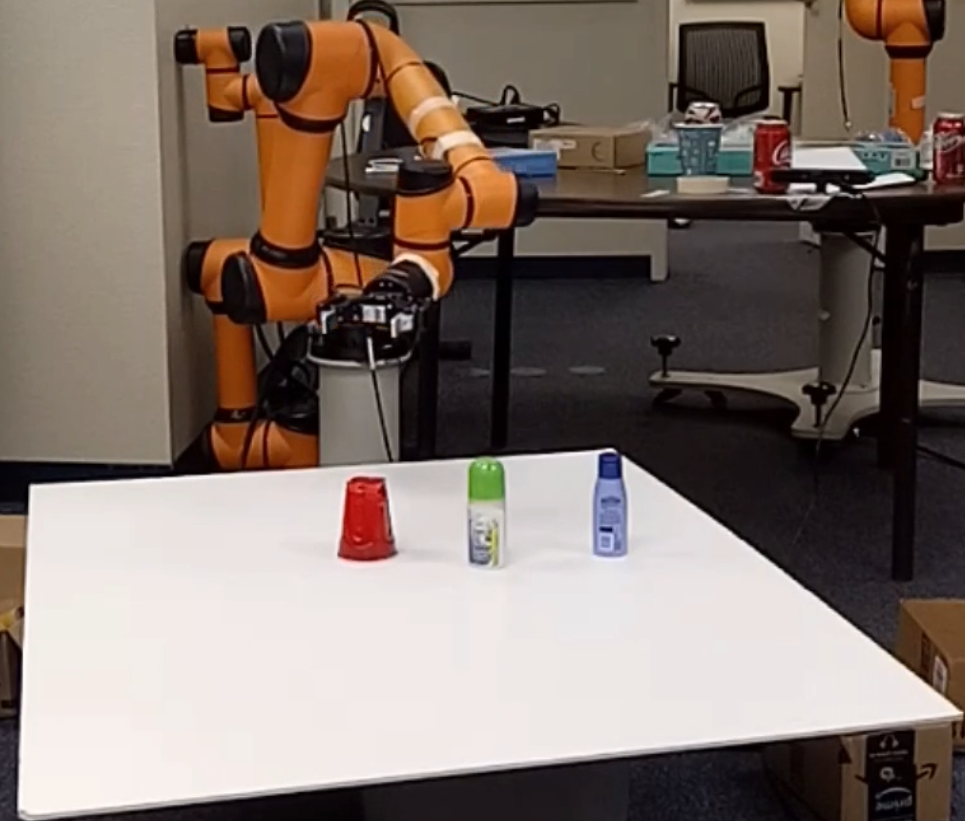}
    \end{subfigure}
    \hspace{\fill}
    \begin{subfigure}[b]{0.2\textwidth}   
        \centering 
        \includegraphics[height=2.5cm]{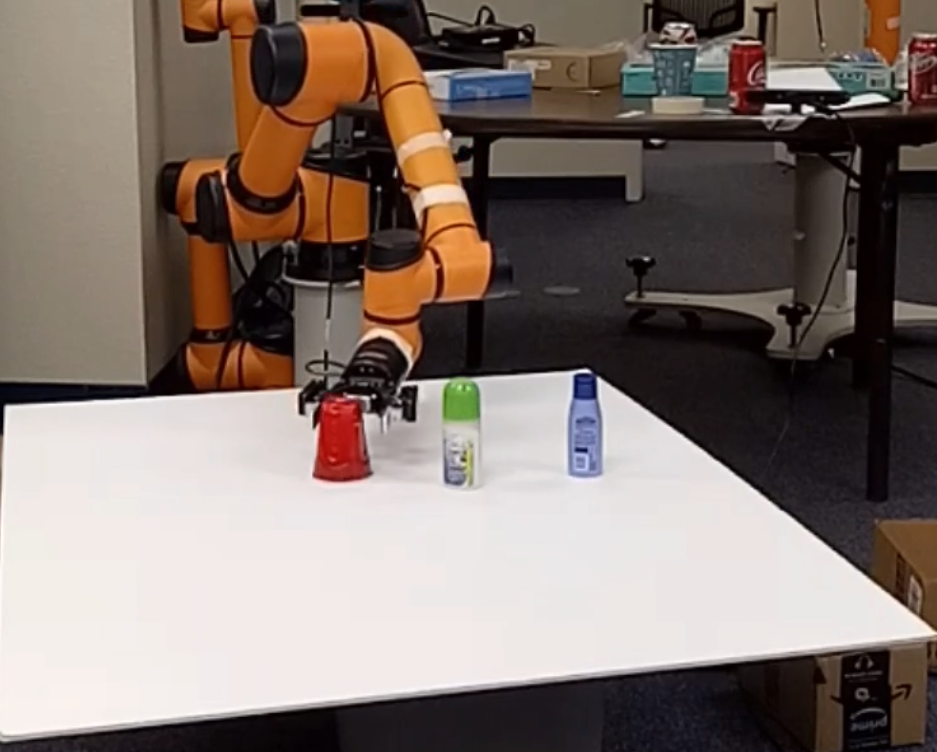}
    \end{subfigure}
	\caption{Generated trajectory replayed on the robot. The trajectory consists of a set of 6-joint positions, which is then replayed on the real robot at 10Hz for safety. This frequency can be up to several hundred Hz as specified by the robot manual.}
    \label{fig:replay_robot}
\end{figure}

\subsection{Camera Calibration}
As grasping objects require accurate positional information, it is essential for the camera to be properly calibrated. There are two different calibration procedures for RGB and depth images that have been performed.
\begin{itemize}
    \item RGB calibration: We use the procedure mentioned in this ROS tutorial \footnote{http://wiki.ros.org/camera\_calibration/Tutorials/MonocularCalibration}. A checkerboard of size 8x6 is used in this step.
    \item Depth calibration: The camera that we use is notorious for having depth error without calibration. We also suffered the same error of about 5 cm at a distance of 50 cm. It can be seen in Figure \ref{fig:depth-error}, the tag from point cloud data is moved 5 cm in front of the actual position. We follow the instructions for calibrating the depth sensor by using jsk\_pcl\_ros package \footnote{https://jsk-recognition.readthedocs.io/en/latest/jsk\_pcl\_ros/calibration.html}. The idea is to align the depth estimation from RGB images with the depth calculated from the depth sensor. We used the same size checkerboard with a smaller grid in order to cover the better our working space while performing the calibration. After calibration, the error is reduced to 1 cm.
\end{itemize}

\subsection{Training YOLOv3}
Object image data was collected using online images from 4 categories: lotion bottles, deodorant bottles, cups, and cans. To come up with this list, we need to test the performance of GDP algorithm on these objects to make sure that the algorithm can generate valid grasps poses. Some transparent objects are not effectively detected by laser beams from the camera, due to which, they were not used the experimentation. Additionally, the neural network used in GPD algorithm to generate grasp poses was trained in certain objects, as a result, it might not generalize with novel objects. A total of 800 images for each class were used for training and validation. After that, the data was manually labelled and fine-tuning was performed with the instruction for training with custom objects from a GitHub repository \footnote{https://github.com/AlexeyAB/darknet}. For the selected object dataset, 80\% of the total images for training and 20\% of the images dataset for testing. The default configuration of weights and hyper-parameters has been chosen for training of object classification model (from darknet53.conv.74). The final performance of the trained object classification model is 84.39\% mAP and 0.5 Intersection over Union (IOU). An example of detected objects is shown in Figure \ref{fig:YOLO}. The trained network is used to return the coordinates of rectangles to cover the detected objects. From these bounding boxes, the centroids for target objects are calculated, as discussed in the previous section.
\subsection{Results}

The trajectory generated by \textit{TrajOpt}, given the transformed point cloud data, the robot model, and the desired grasp pose is shown in Figure \ref{fig:planning}. The scene with point cloud data is transformed into simpler meshes for collision checking. After that, the path is generated to transform the current pose of the gripper to the desired grasp pose. At the final step, the gripper is able to reach the desired grasp pose to prepare to grasp the object. The generated trajectory is replayed on the real robot, which is shown in Figure \ref{fig:replay_robot}. The figure shows the accuracy of the generated path where the gripper is able to come close to the surface without collision. The coke can is also in the middle of the grasp, ready to be grasped. The path planning process for all four objects is performed then combined into a single replay. The generated trajectory consists of multiple 6-joint positions of the robot, which is transmitted to the robot every 0.1 seconds. The frequency of 10 Hz was selected to ensure seamless and safe execution of the proposed experimentation. The path planning for grasp pose and practical execution of object grasping on the four objects can be seen in https://www.youtube.com/watch?v=ujorCGl5Ieo. All source code is available on our ARA lab's GitHub https://github.com/aralab-unr/GraspInPointCloud.

\section{Conclusions and Future Work}
This paper presented the development of an autonomous robotic manipulator for sorting application. Two state-of-the-art Deep Conventional Neural Networks were used to process RGB-D data for both object detection and robust grasp pose generation. Combining with \textit{Trajopt} motion planning, it formed a viable solution for an autonomous sorting robot. Moreover, this paper also discussed the design of a grasp filtering, which works as an interface between the existing grasp pose detection algorithm and the variance in sorting robot system. This can ensure that the different algorithms can work robustly with our gripper's configuration.

The proposed system is validated by an experiment utilizing the detection, grasping and sorting of different object types. The experiment results on various objects show that our proposed combination of deep learning-based object detection model, grasp detection \& filtering, and the manipulator control method is able to provide robust and efficient object grasping and sorting of different objects. The proposed approach can be adapted to different types of manipulators, gripper mechanisms and robots.

There are a few drawbacks to our research, which can be improved in the future. The first problem is the processing time due to large amount of data being used by motion planning and Grasp Pose Detection algorithms. Future research should focus towards improving the data representation to reduce the processing data. Another potential improvement would be towards development of grasp pose filter, which is able to work efficiently with relative position change between robot joints and camera.
Moreover, combining more cameras will provide better point cloud data to represent the object, which in turn will improve the Grasp Pose detection in cluttered environment. 

We also plan to extend this work to multi-manipulator collaboration in which both collaborative and distributed control \cite{PhamIROS2017,DangMFI2016,LaIEEE_TSMCS2015,LaIEEE_TC2013,LaRAS2012} and deep reinforcement learning \cite{NguyenICDL-EpiRob19,PhamSSRR2018,MehdiICERA2018,LaIEEE_TCST2015} will be investigated to allow multiple manipulators to work together efficiently while avoiding collision. The multi-agent cooperative control and sensing research in our previous work  \cite{DangIJARS2019,JinAutomatica2019,ConnellIJARS2018,NguyenIEEE_TCNS2018,PhamIEEE_TSMCS2018,NguyenIEEE_TCNS2017,NguyenRCAR2017,FilibertoIJRNC2017,NguyenAlerton2016,LaICRA2010,LaICRA2009,LaIROS2009,LaACC2009} will be utilized.

\bibliography{ref.bib}

\begin{thebibliography}{10}
\providecommand{\url}[1]{#1}
\csname url@samestyle\endcsname
\providecommand{\newblock}{\relax}
\providecommand{\bibinfo}[2]{#2}
\providecommand{\BIBentrySTDinterwordspacing}{\spaceskip=0pt\relax}
\providecommand{\BIBentryALTinterwordstretchfactor}{4}
\providecommand{\BIBentryALTinterwordspacing}{\spaceskip=\fontdimen2\font plus
\BIBentryALTinterwordstretchfactor\fontdimen3\font minus
  \fontdimen4\font\relax}
\providecommand{\BIBforeignlanguage}[2]{{%
\expandafter\ifx\csname l@#1\endcsname\relax
\typeout{** WARNING: IEEEtran.bst: No hyphenation pattern has been}%
\typeout{** loaded for the language `#1'. Using the pattern for}%
\typeout{** the default language instead.}%
\else
\language=\csname l@#1\endcsname
\fi
#2}}
\providecommand{\BIBdecl}{\relax}
\BIBdecl

\bibitem{gupta2012using}
M.~Gupta and G.~S. Sukhatme, ``Using manipulation primitives for brick sorting
  in clutter,'' in \emph{2012 IEEE International Conference on Robotics and
  Automation}.\hskip 1em plus 0.5em minus 0.4em\relax IEEE, 2012, pp.
  3883--3889.

\bibitem{zeng2018robotic}
A.~Zeng, S.~Song, K.-T. Yu, E.~Donlon, F.~R. Hogan, M.~Bauza, D.~Ma, O.~Taylor,
  M.~Liu, E.~Romo \emph{et~al.}, ``Robotic pick-and-place of novel objects in
  clutter with multi-affordance grasping and cross-domain image matching,'' in
  \emph{2018 IEEE International Conference on Robotics and Automation
  (ICRA)}.\hskip 1em plus 0.5em minus 0.4em\relax IEEE, 2018, pp. 1--8.

\bibitem{guerin2018unsupervised}
J.~Gu{\'e}rin, S.~Thiery, E.~Nyiri, and O.~Gibaru, ``Unsupervised robotic
  sorting: Towards autonomous decision making robots,'' \emph{arXiv preprint
  arXiv:1804.04572}, 2018.

\bibitem{mahler2019learning}
J.~Mahler, M.~Matl, V.~Satish, M.~Danielczuk, B.~DeRose, S.~McKinley, and
  K.~Goldberg, ``Learning ambidextrous robot grasping policies,'' \emph{Science
  Robotics}, vol.~4, no.~26, p. eaau4984, 2019.

\bibitem{JinTIE2017}
L.~{Jin}, S.~{Li}, H.~M. {La}, and X.~{Luo}, ``Manipulability optimization of
  redundant manipulators using dynamic neural networks,'' \emph{IEEE
  Transactions on Industrial Electronics}, vol.~64, no.~6, pp. 4710--4720, June
  2017.

\bibitem{NguyenIRC2019}
H.~{Nguyen} and H.~{La}, ``Review of deep reinforcement learning for robot
  manipulation,'' in \emph{2019 Third IEEE International Conference on Robotic
  Computing (IRC)}, Feb 2019, pp. 590--595.

\bibitem{schulman2013finding}
J.~Schulman, J.~Ho, A.~X. Lee, I.~Awwal, H.~Bradlow, and P.~Abbeel, ``Finding
  locally optimal, collision-free trajectories with sequential convex
  optimization.'' in \emph{Robotics: science and systems}, vol.~9, no.~1.\hskip
  1em plus 0.5em minus 0.4em\relax Citeseer, 2013, pp. 1--10.

\bibitem{SehgalIRC2019}
A.~{Sehgal}, H.~{La}, S.~{Louis}, and H.~{Nguyen}, ``Deep reinforcement
  learning using genetic algorithm for parameter optimization,'' in \emph{2019
  Third IEEE International Conference on Robotic Computing (IRC)}, Feb 2019,
  pp. 596--601.

\bibitem{redmon2018yolov3}
J.~Redmon and A.~Farhadi, ``Yolov3: An incremental improvement,'' \emph{arXiv
  preprint arXiv:1804.02767}, 2018.

\bibitem{liu2016ssd}
W.~Liu, D.~Anguelov, D.~Erhan, C.~Szegedy, S.~Reed, C.-Y. Fu, and A.~C. Berg,
  ``Ssd: Single shot multibox detector,'' in \emph{European conference on
  computer vision}.\hskip 1em plus 0.5em minus 0.4em\relax Springer, 2016, pp.
  21--37.

\bibitem{simonyan2014very}
K.~Simonyan and A.~Zisserman, ``Very deep convolutional networks for
  large-scale image recognition,'' \emph{arXiv preprint arXiv:1409.1556}, 2014.

\bibitem{NguyenICDL-EpiRob19}
H.~{Nguyen}, H.~M. {La}, and M.~{Deans}, ``Hindsight experience replay with
  experience ranking,'' in \emph{2019 Joint IEEE 9th International Conference
  on Development and Learning and Epigenetic Robotics (ICDL-EpiRob)}, Aug 2019,
  pp. 1--6.

\bibitem{redmon2015real}
J.~Redmon and A.~Angelova, ``Real-time grasp detection using convolutional
  neural networks,'' in \emph{2015 IEEE International Conference on Robotics
  and Automation (ICRA)}.\hskip 1em plus 0.5em minus 0.4em\relax IEEE, 2015,
  pp. 1316--1322.

\bibitem{lenz2015deep}
I.~Lenz, H.~Lee, and A.~Saxena, ``Deep learning for detecting robotic grasps,''
  \emph{The International Journal of Robotics Research}, vol.~34, no. 4-5, pp.
  705--724, 2015.

\bibitem{ten2018using}
A.~ten Pas and R.~Platt, ``Using geometry to detect grasp poses in 3d point
  clouds,'' in \emph{Robotics Research}.\hskip 1em plus 0.5em minus 0.4em\relax
  Springer, 2018, pp. 307--324.

\bibitem{gualtieri2016high}
M.~Gualtieri, A.~ten Pas, K.~Saenko, and R.~Platt, ``High precision grasp pose
  detection in dense clutter,'' in \emph{Intelligent Robots and Systems (IROS),
  2016 IEEE/RSJ International Conference on}.\hskip 1em plus 0.5em minus
  0.4em\relax IEEE, 2016, pp. 598--605.

\bibitem{garrido2014automatic}
S.~Garrido-Jurado, R.~Mu{\~n}oz-Salinas, F.~J. Madrid-Cuevas, and M.~J.
  Mar{\'\i}n-Jim{\'e}nez, ``Automatic generation and detection of highly
  reliable fiducial markers under occlusion,'' \emph{Pattern Recognition},
  vol.~47, no.~6, pp. 2280--2292, 2014.

\bibitem{diankov2008openrave}
R.~Diankov and J.~Kuffner, ``Openrave: A planning architecture for autonomous
  robotics,'' \emph{Robotics Institute, Pittsburgh, PA, Tech. Rep.
  CMU-RI-TR-08-34}, vol.~79, 2008.

\bibitem{PhamIROS2017}
H.~X. {Pham}, H.~M. {La}, D.~{Feil-Seifer}, and M.~{Deans}, ``A distributed
  control framework for a team of unmanned aerial vehicles for dynamic wildfire
  tracking,'' in \emph{2017 IEEE/RSJ International Conference on Intelligent
  Robots and Systems (IROS)}, Sep. 2017, pp. 6648--6653.

\bibitem{DangMFI2016}
A.~D. {Dang}, H.~M. {La}, and J.~{Horn}, ``Distributed formation control for
  autonomous robots following desired shapes in noisy environment,'' in
  \emph{2016 IEEE International Conference on Multisensor Fusion and
  Integration for Intelligent Systems (MFI)}, Sep. 2016, pp. 285--290.

\bibitem{LaIEEE_TSMCS2015}
H.~M. {La}, W.~{Sheng}, and J.~{Chen}, ``Cooperative and active sensing in
  mobile sensor networks for scalar field mapping,'' \emph{IEEE Transactions on
  Systems, Man, and Cybernetics: Systems}, vol.~45, no.~1, pp. 1--12, Jan 2015.

\bibitem{LaIEEE_TC2013}
H.~M. {La} and W.~{Sheng}, ``Distributed sensor fusion for scalar field mapping
  using mobile sensor networks,'' \emph{IEEE Transactions on Cybernetics},
  vol.~43, no.~2, pp. 766--778, April 2013.

\bibitem{LaRAS2012}
\BIBentryALTinterwordspacing
H.~M. La and W.~Sheng, ``Dynamic target tracking and observing in a mobile
  sensor network,'' \emph{Robotics and Autonomous Systems}, vol.~60, no.~7, pp.
  996 -- 1009, 2012. [Online]. Available:
  \url{http://www.sciencedirect.com/science/article/pii/S0921889012000565}
\BIBentrySTDinterwordspacing

\bibitem{PhamSSRR2018}
H.~X. {Pham}, H.~M. {La}, D.~{Feil-Seifer}, and L.~{Van Nguyen},
  ``Reinforcement learning for autonomous uav navigation using function
  approximation,'' in \emph{2018 IEEE International Symposium on Safety,
  Security, and Rescue Robotics (SSRR)}, Aug 2018, pp. 1--6.

\bibitem{MehdiICERA2018}
M.~Rahimi, S.~Gibb, Y.~Shen, and H.~M. La, ``A comparison of various approaches
  to reinforcement learning algorithms for multi-robot box pushing,'' in
  \emph{Advances in Engineering Research and Application}, H.~Fujita, D.~C.
  Nguyen, N.~P. Vu, T.~L. Banh, and H.~H. Puta, Eds.\hskip 1em plus 0.5em minus
  0.4em\relax Cham: Springer International Publishing, 2019, pp. 16--30.

\bibitem{LaIEEE_TCST2015}
H.~M. {La}, R.~{Lim}, and W.~{Sheng}, ``Multirobot cooperative learning for
  predator avoidance,'' \emph{IEEE Transactions on Control Systems Technology},
  vol.~23, no.~1, pp. 52--63, Jan 2015.

\bibitem{DangIJARS2019}
\BIBentryALTinterwordspacing
A.~D. Dang, H.~M. La, T.~Nguyen, and J.~Horn, ``Formation control for
  autonomous robots with collision and obstacle avoidance using a rotational
  and repulsive force based approach,'' \emph{International Journal of Advanced
  Robotic Systems}, vol.~16, no.~3, p. 1729881419847897, 2019. [Online].
  Available: \url{https://doi.org/10.1177/1729881419847897}
\BIBentrySTDinterwordspacing

\bibitem{JinAutomatica2019}
\BIBentryALTinterwordspacing
L.~Jin, S.~Li, H.~M. La, X.~Zhang, and B.~Hu, ``Dynamic task allocation in
  multi-robot coordination for moving target tracking: A distributed
  approach,'' \emph{Automatica}, vol. 100, pp. 75 -- 81, 2019. [Online].
  Available:
  \url{http://www.sciencedirect.com/science/article/pii/S0005109818305338}
\BIBentrySTDinterwordspacing

\bibitem{ConnellIJARS2018}
\BIBentryALTinterwordspacing
D.~Connell and H.~M. La, ``Extended rapidly exploring random tree based dynamic
  path planning and replanning for mobile robots,'' \emph{International Journal
  of Advanced Robotic Systems}, vol.~15, no.~3, p. 1729881418773874, 2018.
  [Online]. Available: \url{https://doi.org/10.1177/1729881418773874}
\BIBentrySTDinterwordspacing

\bibitem{NguyenIEEE_TCNS2018}
M.~T. {Nguyen}, H.~M. {La}, and K.~A. {Teague}, ``Collaborative and compressed
  mobile sensing for data collection in distributed robotic networks,''
  \emph{IEEE Transactions on Control of Network Systems}, vol.~5, no.~4, pp.
  1729--1740, Dec 2018.

\bibitem{PhamIEEE_TSMCS2018}
H.~X. {Pham}, H.~M. {La}, D.~{Feil-Seifer}, and M.~C. {Deans}, ``A distributed
  control framework of multiple unmanned aerial vehicles for dynamic wildfire
  tracking,'' \emph{IEEE Transactions on Systems, Man, and Cybernetics:
  Systems}, pp. 1--12, 2018.

\bibitem{NguyenIEEE_TCNS2017}
T.~{Nguyen}, H.~M. {La}, T.~D. {Le}, and M.~{Jafari}, ``Formation control and
  obstacle avoidance of multiple rectangular agents with limited communication
  ranges,'' \emph{IEEE Transactions on Control of Network Systems}, vol.~4,
  no.~4, pp. 680--691, Dec 2017.

\bibitem{NguyenRCAR2017}
T.~{Nguyen} and H.~M. {La}, ``Distributed formation control of nonholonomic
  mobile robots by bounded feedback in the presence of obstacles,'' in
  \emph{2017 IEEE International Conference on Real-time Computing and Robotics
  (RCAR)}, July 2017, pp. 206--211.

\bibitem{FilibertoIJRNC2017}
\BIBentryALTinterwordspacing
F.~MuÃ±oz, E.~S. EspinozaÂ Quesada, H.~M. La, S.~Salazar, S.~Commuri, and
  L.~R. GarciaÂ Carrillo, ``Adaptive consensus algorithms for real-time
  operation of multi-agent systems affected by switching network events,''
  \emph{International Journal of Robust and Nonlinear Control}, vol.~27, no.~9,
  pp. 1566--1588, 2017. [Online]. Available:
  \url{https://onlinelibrary.wiley.com/doi/abs/10.1002/rnc.3687}
\BIBentrySTDinterwordspacing

\bibitem{NguyenAlerton2016}
T.~{Nguyen}, T.~{Han}, and H.~M. {La}, ``Distributed flocking control of mobile
  robots by bounded feedback,'' in \emph{2016 54th Annual Allerton Conference
  on Communication, Control, and Computing (Allerton)}, Sep. 2016, pp.
  563--568.

\bibitem{LaICRA2010}
H.~M. {La} and W.~{Sheng}, ``Flocking control of multiple agents in noisy
  environments,'' in \emph{2010 IEEE International Conference on Robotics and
  Automation}, May 2010, pp. 4964--4969.

\bibitem{LaICRA2009}
------, ``Flocking control of a mobile sensor network to track and observe a
  moving target,'' in \emph{2009 IEEE International Conference on Robotics and
  Automation}, May 2009, pp. 3129--3134.

\bibitem{LaIROS2009}
------, ``Adaptive flocking control for dynamic target tracking in mobile
  sensor networks,'' in \emph{2009 IEEE/RSJ International Conference on
  Intelligent Robots and Systems}, Oct 2009, pp. 4843--4848.

\bibitem{LaACC2009}
H.~M. {La} and {Weihua Sheng}, ``Moving targets tracking and observing in a
  distributed mobile sensor network,'' in \emph{2009 American Control
  Conference}, June 2009, pp. 3319--3324.

\end{thebibliography}
\bibliographystyle{IEEEtran}
\end{document}